\documentclass[letterpaper]{article}

\usepackage[draft]{format}
\usepackage{times}
\usepackage{helvet}
\usepackage{courier}
\usepackage[hyphens]{url}
\usepackage{graphicx}
\usepackage{natbib}
\usepackage{amsmath,amssymb}
\usepackage{booktabs}

\graphicspath{{./}{figures/}}

\title{DriftingMol: Decoder-Coupled Drift for One-Pass Property-Conditional Molecular Generation}
\author{Jiangjie Qiu, Yijun Li, Wentao Li, Xiaonan Wang\thanks{Corresponding author.}}
\affiliations{Beijing Key Laboratory of Artificial Intelligence for Advanced Chemical Engineering Materials}

\begin{document}
\maketitle

\begin{abstract}
Property-conditional molecular generation should produce valid, diverse
molecules while responding to continuous target values at low sampling cost.
We introduce DriftingMol, a two-stage framework that adapts drifting models to
a SELFIES latent molecular space. A frozen SELFIES $\beta$-VAE provides the
latent space, and the hidden representation of its decoder serves as the drift
feature map. In decoder-coupled drift, decoder weights remain fixed, but drift
gradients are backpropagated through the decoder feature map to a DiT generator,
inducing a pullback metric aligned with molecular decoding. On ZINC250K, the
default setting achieves QED Spearman correlation 0.493 with 94.7\% uniqueness,
while the strongest decoder-coupled condition reaches 0.510. Under
protocol-matched
four-property conditioning, decoder-coupled drift reaches mean Spearman
correlation up to 0.598. Across 15 controlled variants, models that preserve
the gradient path through decoder features consistently outperform
latent-space, random-feature, and external-feature drift variants, while
detached or stop-gradient decoder controls yield near-zero QED correlation and
very low uniqueness. These results support decoder-coupled drift as an
efficient mechanism for property-biased molecular generation with one generator
evaluation plus one frozen decoder pass.
\end{abstract}

\section{Introduction}

De novo molecular design asks for novel molecules that satisfy desired
physicochemical or drug-likeness constraints. The setting is difficult because
validity is discrete, property control is continuous, and strong generative
models often require iterative sampling. SMILES and graph generators have
produced major progress through VAEs, flows, reinforcement learning, and
diffusion models \cite{gomez2018automatic,jtvae,moflow,digress,gdss,
olivecrona2017molecular}. However, iterative methods can be expensive at
inference, while latent or graph decoders can lose continuous property signal
when discrete decoding collapses nearby latent states.

SELFIES provides a validity-preserving molecular string representation: every
SELFIES string maps to a valid molecule
\cite{krenn2020selfies,krenn2022selfies}. We use this property deliberately to
separate validity from controllability. A SELFIES $\beta$-VAE supplies a
regularized continuous latent space, and validity is inherited from decoding
rather than learned by the conditional generator. We then train a one-pass
generator using the drifting model objective \cite{deng2026drifting}, which
learns a noise-to-data map without iterative denoising.

The central question is which feature space should define the molecular drift
field. The original drifting formulation uses an external feature extractor.
For molecules, we find that feature quality alone is insufficient. Instead, the
frozen decoder already contains a differentiable structural representation of
how latent vectors become token sequences. DriftingMol reuses the decoder
hidden state as $\varphi$ and keeps the gradient path
$\varphi(\hat{z}) \to \hat{z} \to \theta$ open. The decoder weights are frozen,
but its Jacobian supplies an adaptive geometry for generator updates.

Our contributions are:
\begin{itemize}
    \item We introduce a SELFIES latent drifting pipeline for low-cost
    property-conditional molecular generation.
    \item We propose decoder-coupled drift, which uses the frozen VAE decoder as
    the drift feature space while preserving gradients through the decoder to
    the generator.
    \item We provide a 15-condition ablation showing that decoder coupling is the
    main mechanism behind QED control, and that external trained features do
    not replace this gradient coupling.
    \item We extend the mechanism study to protocol-matched four-property conditioning,
    z-diversity sensitivity, implementation checks, same-backbone generator
    baselines, and a separate graph-route stress test, while keeping inference
    to one generator forward pass plus one frozen decoder pass.
\end{itemize}

\section{Related Work}

\paragraph{Molecular generation.}
Molecular generators include VAEs \cite{jtvae,graphvae}, continuous-latent
SMILES models \cite{gomez2018automatic}, flows \cite{moflow,graphnvp}, graph
diffusion models \cite{digress,gdss}, and reinforcement-learning or
optimization systems
\cite{olivecrona2017molecular,zhou2019optimization,eckmann2022limo,
yang2021freed,lee2023mood}. These methods differ in whether they optimize
properties, target prescribed property values, or match a data distribution.
DriftingMol focuses on amortized property targeting with a single generator
evaluation followed by decoding.

\paragraph{Drifting and guidance.}
Drifting models train a generator with a kernel drift field between generated
and reference samples \cite{deng2026drifting}. They are related in spirit to
flow matching \cite{lipman2023flow}, but avoid test-time ODE integration.
Classifier-free guidance (CFG) steers conditional generators by mixing
conditional and unconditional information \cite{ho2022classifierfree}. We embed the
guidance scale into training so inference does not require a second generator
evaluation for conditional/unconditional mixing.

\paragraph{SELFIES representations.}
SELFIES provides representation-level syntactic and valence validity for
decoded strings \cite{krenn2020selfies}. In DriftingMol, final validity is
still evaluated after decoding and RDKit canonicalization, but this
representation property makes validity largely non-discriminative across model
conditions, allowing the ablation to focus on conditional control, uniqueness,
novelty, and diversity.

\section{Method}

\subsection{Two-Stage Pipeline}

Stage 1 trains a SELFIES $\beta$-VAE on ZINC250K \cite{irwin2012zinc}. The
encoder maps token sequences to $z \in \mathbb{R}^{256}$; the one-shot
Transformer decoder predicts all SELFIES positions in parallel. The VAE loss is
\begin{equation}
    \mathcal{L}_{\mathrm{VAE}} =
    \mathcal{L}_{\mathrm{recon}} +
    \beta D_{\mathrm{KL}}(q(z|x)\|\mathcal{N}(0,I)),
\end{equation}
with $\beta=0.01$. The trained VAE reaches 82.3\% exact reconstruction and
99.1\% token accuracy on held-out molecules.

Stage 2 freezes the VAE and trains a DiT-style generator
\cite{peebles2023scalable}
$f_\theta(\epsilon,c,\alpha)$ with 20.2M parameters. It maps Gaussian noise,
optional property condition $c$, and embedded guidance scale $\alpha$ to a
latent $\hat{z}$ in one generator forward pass. Decoding $\hat{z}$ through the
frozen SELFIES decoder yields the final molecule.

\subsection{Coupled Decoder Drift}

Given generated latent $\hat{z}_i$ and reference latents $z_j^{\mathrm{ref}}$,
we extract decoder hidden features
$\varphi(\hat{z}_i),\varphi(z_j^{\mathrm{ref}}) \in \mathbb{R}^{512}$. The
drift field at temperature $\tau$ is
\begin{equation}
    V_{\tau,i} =
    \sum_j W^\tau_{ij}
    \left(\varphi(z_j^{\mathrm{ref}})-\varphi(\hat{z}_i)\right),
\end{equation}
where
\begin{equation}
    \begin{aligned}
    A^\tau &= \operatorname{softmax}_{\mathrm{row}}(S/\tau),\\
    B^\tau &= \operatorname{softmax}_{\mathrm{col}}(S/\tau),\\
    W^\tau_{ij} &= (A^\tau_{ij}B^\tau_{ij})^{1/2}.
    \end{aligned}
\end{equation}
and $S_{ij}=-\|\varphi(\hat{z}_i)-\varphi(z_j^{\mathrm{ref}})\|/d_{\mathrm{global}}$.
Thus the row- and column-normalized kernels are combined elementwise. The scale
$d_{\mathrm{global}}$ is the mean pairwise feature distance within the current
drift group, computed from positive references, generated-sample negatives
with self-distances excluded, and unconditional negatives when CFG is active.
For the default DriftingMol model, $\tau \in \{0.5,1.0,2.0\}$ and each
temperature is normalized by a precomputed scale $\lambda_\tau$. With fixed
normalization, $\lambda_\tau$ is computed before Stage 2 on cached training
features using the same generated/reference group sizes as training:
$\lambda_\tau=(\mathbb{E}_i\|V_{\tau,i}^{\mathrm{cal}}\|_2^2/D)^{1/2}$, where
$D$ is the drift-feature dimension, and is then kept fixed.

The drift loss is
\begin{equation}
    \mathcal{L}_{\mathrm{drift}} =
    \left\|\varphi(\hat{z}) -
    \operatorname{sg}\left(\varphi(\hat{z})+\sum_\tau V_\tau/\lambda_\tau\right)
    \right\|_2^2.
\end{equation}
Only the target is stop-gradient. The decoder parameters are frozen, but the
left-hand $\varphi(\hat{z})$ retains gradients, giving
\begin{equation}
    \frac{\partial \mathcal{L}_{\mathrm{drift}}}{\partial \theta}
    =
    \frac{\partial \mathcal{L}_{\mathrm{drift}}}{\partial \varphi}
    \frac{\partial \varphi}{\partial \hat{z}}
    \frac{\partial \hat{z}}{\partial \theta}.
\end{equation}
This is the coupling absent from random, external, detached, and z-only
feature-space controls.

\subsection{Diversity and Guidance}

We add a kNN latent repulsion term to prevent mode collapse:
\begin{equation}
    \mathcal{L}_{\mathrm{zdiv}} =
    \frac{1}{K}\sum_{k=1}^{K}
    \max(0,m-\|z_i-z_{\mathrm{nn}_k(i)}\|),
\end{equation}
with $K=5$ and margin $m=3.0$. The total loss is
\begin{equation}
    \mathcal{L}=\mathcal{L}_{\mathrm{drift}}+
    \lambda_z\mathcal{L}_{\mathrm{zdiv}},
\end{equation}
with default $\lambda_z=2.0$ unless otherwise stated.

\begin{figure*}[t]
\centering
\includegraphics[width=\textwidth,height=0.66\textheight,keepaspectratio]{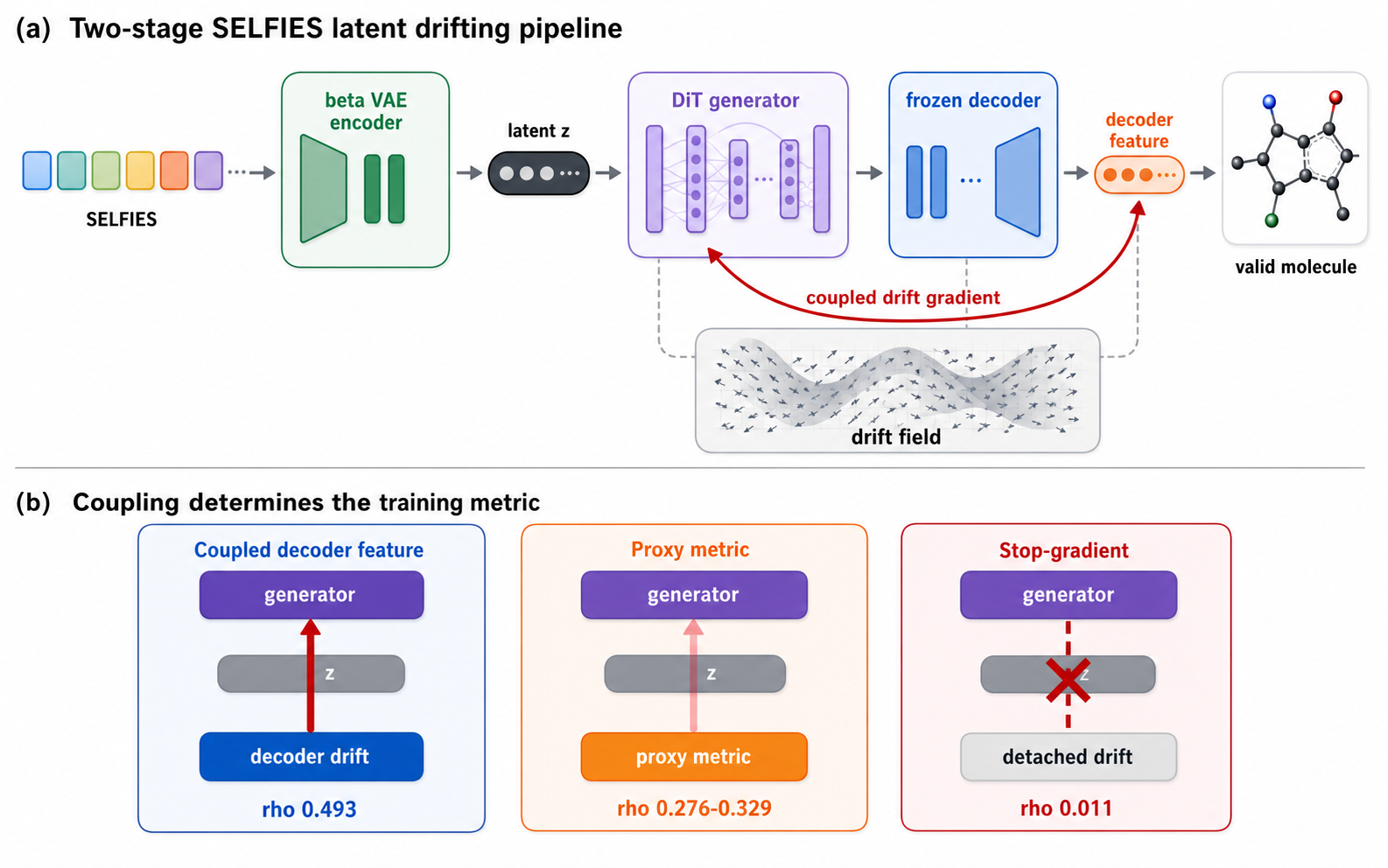}
\caption{DriftingMol overview. (a) SELFIES molecules are encoded once into a
latent cache, then a conditional DiT predicts a drifted latent in one generator
pass. The frozen decoder hidden state defines the drift feature space used by
the kernel drift objective. (b) Coupling the drift loss to decoder-$\varphi$
keeps the feature metric differentiable with respect to the generated latent,
whereas proxy metrics or stop-gradient controls weaken the training signal.}
\label{fig:main}
\end{figure*}

For scalar QED experiments with property bins, each target group samples
$N_{\mathrm{pos}}$ positive references from the matched bin. In hybrid mode,
half are random same-bin references and half are nearest neighbors in cached
decoder-feature space to the generated group mean; property-only controls use
nearest property values instead. In the v2 four-property protocol, positives
are selected by property-space kNN without bins. CFG-like repulsion samples
$N_{\mathrm{unc}}$ unconditional references from the training cache. With
$N_g$ generated samples in a group, the unconditional weight is
\begin{equation}
    w(\alpha)=\max\left\{0,\frac{(\alpha-1)(N_g-1)}{N_{\mathrm{unc}}}\right\}.
\end{equation}
For $\alpha>1$, unconditional-negative logits are
\begin{equation}
    \begin{aligned}
    \ell^{\mathrm{unc}}_{ik}
    &= -\frac{\|\varphi(\hat z_i)-\varphi(z^{\mathrm{unc}}_k)\|}
    {\tau d_{\mathrm{global}}}\\
    &\quad+\log w(\alpha)
    \end{aligned}
\end{equation}
before the same row/column normalization as above. For $\alpha=1$, $w=0$ and
the unconditional-negative branch is omitted rather than evaluated as
$\log 0$. The guidance scale is sampled during training from
$p(\alpha)\propto\alpha^{-3}$ on $[1,4]$ and embedded into the generator. At
test time we evaluate $\alpha \in \{1.0,1.5,2.0,3.0,5.0\}$ with one generator
pass per sample.
Figure~\ref{fig:main} summarizes the two-stage pipeline and the coupling
contrast tested in the ablations.

\section{Mathematical Analysis}

This section formalizes the mechanism tested by the ablations. Let
$g_\theta(\epsilon,c,\alpha)=\hat z$ denote the DiT generator and let
$\varphi:\mathbb{R}^{256}\rightarrow\mathbb{R}^{512}$ denote the hidden feature
map of the frozen SELFIES decoder.  The decoder weights are fixed throughout
Stage 2, but $\varphi(\hat z)$ is still a differentiable function of the
generated latent.  For a mini-batch sample, write
\begin{equation}
    y_\theta = \varphi(g_\theta(\epsilon,c,\alpha)).
\end{equation}
The drift target used by DriftingMol can be written abstractly as
\begin{equation}
    b_\theta = \operatorname{sg}(y_\theta + v_\theta),
\end{equation}
where $v_\theta=\sum_\tau V_\tau/\lambda_\tau$ is the kernel drift vector in
decoder-feature space and $\operatorname{sg}$ blocks target-side gradients.
The per-sample loss is
\begin{equation}
    \ell(\theta)=\frac{1}{2}\|y_\theta-b_\theta\|_2^2 .
\end{equation}

\paragraph{Proposition 1: decoder-coupled drift supplies a decoder pullback
gradient.}
Assume $g_\theta$ and $\varphi$ are differentiable at the generated sample.
With the target side stopped as above,
\begin{equation}
    \nabla_\theta \ell
    =
    J_{g_\theta}^{\top}
    J_{\varphi}^{\top}
    (y_\theta-b_\theta)
    =
    -J_{g_\theta}^{\top}J_{\varphi}^{\top}v_\theta ,
    \label{eq:coupled-gradient}
\end{equation}
where $J_{g_\theta}=\partial g_\theta/\partial\theta$ and
$J_\varphi=\partial\varphi/\partial z$ at $z=\hat z$.

\emph{Proof.}
Since $b_\theta$ is stop-gradient, it is treated as a constant by
backpropagation.  Differentiating
$\ell=\frac{1}{2}(y_\theta-b_\theta)^\top(y_\theta-b_\theta)$ gives
\begin{equation}
    \nabla_\theta \ell
    =
    \left(\frac{\partial y_\theta}{\partial\theta}\right)^\top
    (y_\theta-b_\theta).
\end{equation}
The chain rule yields
$\partial y_\theta/\partial\theta=J_\varphi J_{g_\theta}$, and substituting
$b_\theta=\operatorname{sg}(y_\theta+v_\theta)$ gives
$y_\theta-b_\theta=-v_\theta$ in the backward pass. This proves
Eq.~\eqref{eq:coupled-gradient}. \hfill $\square$

The proposition explains the qualitative separation in Table~\ref{tab:qed-main}.
If the left-hand feature $\varphi(\hat z)$ is detached, then
$\partial y_\theta/\partial\theta=0$ and the drift loss cannot update the
generator.  If drift is applied directly in $z$-space, the update is
$-J_{g_\theta}^{\top}v_z$ and the decoder Jacobian is absent.  If an external
feature map $\psi$ is used, the update becomes
$-J_{g_\theta}^{\top}J_\psi^\top v_\psi$, which follows the geometry of
$\psi$ rather than the geometry of the frozen molecular decoder.

\paragraph{Proposition 2: decoder features define the local metric optimized by
the generator.}
For a small latent perturbation $\Delta z$, decoder-feature distance satisfies
\begin{equation}
    \|\varphi(z+\Delta z)-\varphi(z)\|_2^2
    =
    \Delta z^\top G_\varphi(z)\Delta z
    + o(\|\Delta z\|^2),
    \label{eq:pullback}
\end{equation}
where
\begin{equation}
    G_\varphi(z)=J_\varphi(z)^\top J_\varphi(z)
\end{equation}
is a positive semidefinite pullback metric on latent space.

\emph{Proof.}
By first-order Taylor expansion,
$\varphi(z+\Delta z)=\varphi(z)+J_\varphi(z)\Delta z+o(\|\Delta z\|)$.
Taking the squared Euclidean norm of the feature difference gives
\begin{align}
&\|J_\varphi(z)\Delta z+o(\|\Delta z\|)\|_2^2 \nonumber\\
&\quad =
\Delta z^\top J_\varphi(z)^\top J_\varphi(z)\Delta z
+ o(\|\Delta z\|^2).
\end{align}
The matrix $J_\varphi^\top J_\varphi$ is positive semidefinite because
$u^\top J_\varphi^\top J_\varphi u=\|J_\varphi u\|_2^2\ge0$ for any $u$.
\hfill $\square$

Thus decoder-coupled drift does more than choose a representation.  It changes
the metric in which the generator is trained.  The update in
Eq.~\eqref{eq:coupled-gradient} is the feature-space drift vector transported
back through the decoder Jacobian.  This weights latent directions according to
how strongly the frozen decoder changes its molecular hidden state along those
directions.  A trained property encoder may be predictive, but unless its
Jacobian matches $J_\varphi$, it optimizes a different pullback metric.

\paragraph{Proposition 3: the kernel drift target is a bounded attraction field
under bounded decoder features.}
Assume $\|\varphi(z)\|_2\le R$ for generated and reference latents in the
mini-batch.  For
\begin{equation}
    V_{\tau,i}=\sum_j W_{ij}^{\tau}
    \left(\varphi(z_j^{\mathrm{ref}})-\varphi(\hat z_i)\right),
\end{equation}
with $W_{ij}^{\tau}\ge0$, the drift magnitude obeys
\begin{equation}
    \|V_{\tau,i}\|_2
    \le
    2R\sum_j W_{ij}^{\tau}.
\end{equation}

\emph{Proof.}
The row and column softmax factors are nonnegative, so their geometric mean
$W_{ij}^{\tau}$ is nonnegative.  By the triangle inequality,
\begin{align}
\|V_{\tau,i}\|_2
&\le
\sum_j W_{ij}^{\tau}
\|\varphi(z_j^{\mathrm{ref}})-\varphi(\hat z_i)\|_2 \nonumber\\
&\le
\sum_j W_{ij}^{\tau}
\left(\|\varphi(z_j^{\mathrm{ref}})\|_2
+\|\varphi(\hat z_i)\|_2\right) \nonumber\\
&\le 2R\sum_j W_{ij}^{\tau}.
\end{align}
\hfill $\square$

This bound is simple but useful: the drift target is a nonnegative combination
of reference-to-generated feature differences, so temperature and normalization
control the attraction strength without changing the differentiable metric in
Eq.~\eqref{eq:pullback}.

\paragraph{Proposition 4: conditional drift is a feature-space mean shift.}
For fixed temperature $\tau$ and generated sample $i$, define
$c_i^\tau=\sum_j W_{ij}^\tau$.  If $c_i^\tau>0$, define the local conditional
feature barycenter
\begin{equation}
    \mu_i^\tau =
    \frac{1}{c_i^\tau}
    \sum_j W_{ij}^\tau\varphi(z_j^{\mathrm{ref}}).
\end{equation}
Then
\begin{equation}
    V_{\tau,i}
    =
    c_i^\tau\left(\mu_i^\tau-\varphi(\hat z_i)\right).
    \label{eq:mean-shift}
\end{equation}

\emph{Proof.}
Expanding the definition of $V_{\tau,i}$,
\begin{align}
V_{\tau,i}
&=\sum_j W_{ij}^\tau\varphi(z_j^{\mathrm{ref}})
-\left(\sum_j W_{ij}^\tau\right)\varphi(\hat z_i) \nonumber\\
&=c_i^\tau\mu_i^\tau-c_i^\tau\varphi(\hat z_i).
\end{align}
\hfill $\square$

Thus the property-conditioned reference sampler changes the barycenter
$\mu_i^\tau$ toward target-compatible molecules, while the decoder pullback in
Eq.~\eqref{eq:coupled-gradient} determines how that feature-space displacement
updates the latent generator.  This separates two design roles: conditioning
selects the reference distribution, and coupling selects the geometry used to
move generated latents toward it.

\paragraph{Proposition 5: the z-diversity term repels collapsed neighbors.}
Consider one active hinge term
\begin{equation}
    h(z_i,z_j)=\max(0,m-\|z_i-z_j\|_2).
\end{equation}
When $0<\|z_i-z_j\|_2<m$,
\begin{equation}
    \nabla_{z_i} h =
    -\frac{z_i-z_j}{\|z_i-z_j\|_2},
    \qquad
    \nabla_{z_j} h =
    -\frac{z_j-z_i}{\|z_i-z_j\|_2}.
\end{equation}
Therefore a gradient-descent step on $h$ moves the pair away from each other.

\emph{Proof.}
In the active region, $h=m-\|z_i-z_j\|_2$.  Differentiating the Euclidean norm
gives
\begin{equation}
    \nabla_{z_i}\|z_i-z_j\|_2 =
    \frac{z_i-z_j}{\|z_i-z_j\|_2},
\end{equation}
and the expression for $\nabla_{z_i}h$ follows by the negative sign.  The
$z_j$ derivative is symmetric.  A gradient-descent update
$z_i\leftarrow z_i-\eta\nabla_{z_i}h$ adds
$\eta(z_i-z_j)/\|z_i-z_j\|_2$ to $z_i$, increasing the pairwise distance to
first order. \hfill $\square$

This explains why the diversity-free decoder-coupling ablation can achieve high correlation but
lower uniqueness:
without the repulsive term, the drift objective may place many noise samples
near the same attractive feature basin.  With z-diversity, nearby generated
latents receive an explicit counter-force while retaining the decoder-coupled
property-control gradient.

\paragraph{Proposition 6: representation-level validity and inference cost are
inherited from the two-stage construction.}
Let $\mathcal{S}$ be the SELFIES alphabet and let
$T:\mathcal{S}^{*}\rightarrow\mathcal{M}_{\mathrm{valid}}$ be the SELFIES
semantic map from token strings to valid molecular graphs.  If the frozen
decoder outputs a sequence $\hat s\in\mathcal{S}^{*}$, then the generated
molecule $T(\hat s)$ is valid at the representation level.  Moreover, for a fixed condition
$(c,\alpha)$, generation requires one evaluation of $g_\theta$ and one VAE
decoder pass.

\emph{Proof.}
The generator produces a continuous latent $\hat z=g_\theta(\epsilon,c,\alpha)$.
The frozen decoder maps $\hat z$ to SELFIES tokens $\hat s$.  By the defining
property of SELFIES, every token string in $\mathcal{S}^{*}$ maps through $T$
to a syntactically valid molecular graph, so $T(\hat s)$ is valid at the
representation level.  The
computational graph at inference contains no denoising chain, ODE solver, or
iterative refinement: $\hat z$ is produced by one forward call to $g_\theta$,
then decoded once. \hfill $\square$

Together, these statements identify what is being tested experimentally.
SELFIES validity is inherited from the representation, one-pass generator inference follows
from the generator architecture, and the nontrivial empirical question is
whether the decoder pullback metric in Eq.~\eqref{eq:pullback} improves
property control without reducing diversity.  The mean-shift and z-diversity
propositions explain why the ablation must evaluate both correlation and
uniqueness rather than property correlation alone.  The ablations in the next
section are designed exactly around this question.

\section{Experiments}

\paragraph{Dataset and metrics.}
We evaluate on ZINC250K \cite{irwin2012zinc} with QED, synthetic accessibility
(SA), LogP, and molecular weight targets. We report validity (V), uniqueness
(U), novelty (N), Spearman correlation $\rho$, MAE, slope, internal diversity
(IntDiv), and scaffold diversity. Compact tables focus
on $\rho$, slope, U, and MAE. Since SELFIES decoding supplies representation-level
validity, V is non-discriminative; RDKit canonicalization is still used for
uniqueness, novelty, and scaffold metrics, and failed RDKit parses would count
against validity.

\paragraph{Evaluation protocol.}
We use a fixed latent-cache split of 199,564/24,945/24,946
train/validation/test molecules. Each conditional checkpoint is evaluated on
10,000 generated molecules over a fixed guidance grid
$\alpha\in\{1.0,1.5,2.0,3.0,5.0\}$. Tables report the grid
point with the highest target correlation subject to the pre-specified
feasibility constraints
V$\ge$95\%, U$\ge$10\%, and N$\ge$95\%.
If a failed-control row has no feasible grid point, we retain it as a
diagnostic control and report the highest-correlation grid point, with the
infeasibility visible in the reported V/U/N metrics.
This grid-selected $\alpha$ protocol is used to compare mechanisms under a
shared stress-test contract; for deployment, $\alpha$ should be fixed by a
validation-only selection rule before test evaluation.
Accordingly, we emphasize broad performance tiers, three-seed aggregates for
key QED settings, and ablation consistency rather than fine-grained rank
differences.

\paragraph{Backbone control.}
The stage-1 SELFIES VAE prior is tracked separately from conditional drift:
unconditional prior sampling gives V=100.0\%, U=98.9\%, N=100.0\%, and 4,946
unique molecules among 5,000 samples, but it has no target input and therefore
does not provide prescribed property control. This separates representation
validity from the conditional-control claim.

\paragraph{Matched references and baselines.}
We evaluate two nonparametric QED references under the same 20 target bins and
10,000-sample metric code. Target-bin retrieval reaches $\rho=0.999$ and
MAE=0.008 but has 0\% novelty. A frozen-VAE latent-jitter reference
($\sigma=0.1$ around retrieved training latents) reaches $\rho=0.974$ and
MAE=0.017 but only 13.0\% novelty, showing that near-exact QED matching is easy
when training neighborhoods are reused. Same-backbone conditional latent VAE,
WGAN-GP, DDPM, and Flow-Matching generator baselines remain at
$\rho=0.014\pm0.026$, $0.151\pm0.013$, $0.048\pm0.015$, and
$0.080\pm0.004$ over three seeds. A fixed linear property head gives
$\rho=0.046\pm0.150$. These generator baselines use the same frozen SELFIES
latent cache, train split, scalar QED condition, target bins, sample count, and
metric code; they are diagnostic same-backbone baselines rather than exhaustive
retuned implementations of each model family. Each baseline uses 80 epochs,
250 steps per epoch, batch size 1024, hidden width 512, and three MLP blocks;
DDPM uses 100 diffusion steps, Flow Matching uses 50 Euler steps, and WGAN-GP
uses three critic updates per generator update. All baseline latents are
decoded by the same frozen SELFIES VAE.

\paragraph{Mechanism ablations.}
The 15 ablation conditions isolate drift space, gradient flow, temperature
aggregation, z-diversity, and external feature extractors. The
decoder-coupled family contains the default DriftingMol setting, a
single-temperature setting, and a diversity-free setting. Feature-space
controls include latent-space drift, random-feature drift, and LatentMAE
features; gradient controls include stop-gradient and detached-decoder drift.
Table~\ref{tab:ablation-defs} summarizes the naming convention used in the
main result tables.
LatentMAE denotes an auxiliary masked-autoencoding feature extractor trained on
cached SELFIES-VAE latents, optionally with property-prediction supervision; in
these controls it is frozen and used as an external $\varphi$ rather than as
part of the molecule decoder. Layer-balanced decoder coupling uses the same
decoder feature source as DriftingMol but averages matched decoder layers
before computing the drift field. The v2 no-binning multi-property protocol
samples property targets continuously rather than through QED-style target
bins.

\begin{table*}[t]
\centering
\caption{Ablation definitions. All rows use the same SELFIES VAE latent cache
unless otherwise noted. ``Open'' means gradients from the drift loss pass
through the feature map to the generated latent.}
\label{tab:ablation-defs}
\scriptsize
\resizebox{\textwidth}{!}{%
\begin{tabular}{l l l l l l}
\toprule
Family & Feature map & Gradient path & Drift/reference space & Diversity & Purpose \\
\midrule
DriftingMol & decoder hidden state & open & hybrid positive references, multi-$\tau$ & yes & main decoder-coupled setting \\
Single-$\tau$ & decoder hidden state & open & hybrid references, one temperature & yes & temperature-aggregation control \\
No-diversity & decoder hidden state & open & hybrid references, multi-$\tau$ & no & isolates z-diversity regularization \\
Latent-space / random & latent vector or fixed random map & open & non-decoder feature space & yes & tests whether any feature metric suffices \\
LatentMAE drift & plain trained latent feature map & open & external $\varphi$ drift & yes & tests trained $\varphi$ quality \\
LatentMAE-guided drift & property-supervised latent feature map & open & external property-aware $\varphi$ drift & yes & tests property-aware $\varphi$ \\
LatentMAE + latent drift & plain trained latent feature map & open & external $\varphi$ plus z-space drift & yes & tests added latent geometry \\
LatentMAE + decoder & plain trained latent feature map & open & external $\varphi$ plus decoder auxiliary drift & yes & tests decoder auxiliary signal \\
LatentMAE-guided + decoder & property-supervised latent feature map & open & property-aware $\varphi$ plus decoder auxiliary drift & yes & tests combined external/decoder signal \\
Detached / stop-grad & decoder hidden state & blocked or detached & decoder features without decoder-Jacobian coupling & mixed & tests gradient coupling \\
Property / ridge heads & scalar property predictor & n/a & no drift field & n/a & calibration baselines, not drifting models \\
\bottomrule
\end{tabular}
}
\end{table*}

\subsection{QED Conditional Generation}

\begin{table}[t]
\centering
\caption{QED-conditional generation on ZINC250K. Rows report the best feasible
$\alpha$ on the evaluation grid when available; failed controls with no
feasible grid point are retained as diagnostic rows. Slope measures
target-to-actual calibration. The non-drift property-head baseline is a
calibration baseline rather than a drifting model.}
\label{tab:qed-main}
\resizebox{\columnwidth}{!}{%
\begin{tabular}{l c c c c c}
\toprule
Model / control & $\alpha$ & $\rho$ & Slope & U (\%) & MAE \\
\midrule
Decoder coupling (no diversity) & 5.0 & 0.510 & 0.828 & 74.1 & 0.194 \\
Decoder coupling (single $\tau$) & 5.0 & 0.500 & 0.793 & 94.1 & 0.204 \\
\textbf{DriftingMol} & 5.0 & 0.493 & 0.796 & 94.7 & 0.200 \\
\midrule
Property-head baseline$^\dagger$ & 1.0 & 0.449 & 0.495 & 97.3 & 0.393 \\
\midrule
LatentMAE-guided drift & 5.0 & 0.329 & 0.499 & 94.8 & 0.259 \\
LatentMAE + latent drift & 5.0 & 0.312 & 0.499 & 97.2 & 0.244 \\
LatentMAE drift & 5.0 & 0.297 & 0.472 & 96.3 & 0.249 \\
LatentMAE-guided + decoder & 5.0 & 0.296 & 0.449 & 97.8 & 0.202 \\
Latent-space drift & 5.0 & 0.286 & 0.461 & 98.5 & 0.245 \\
LatentMAE + decoder & 3.0 & 0.281 & 0.432 & 98.8 & 0.200 \\
Random-feature drift control & 5.0 & 0.276 & 0.433 & 97.9 & 0.234 \\
\midrule
Detached decoder-feature control & 3.0 & 0.031 & 0.016 & 0.3 & 0.274 \\
Detached decoder + latent drift & 3.0 & 0.019 & 0.014 & 0.3 & 0.273 \\
Stop-gradient decoder control & 3.0 & 0.011 & 0.010 & 0.3 & 0.272 \\
Ridge-head baseline$^\dagger$ & 3.0 & 0.009 & 0.005 & 0.3 & 0.273 \\
\bottomrule
\end{tabular}
}
\end{table}

\begin{figure*}[t]
\centering
\includegraphics[width=0.94\textwidth]{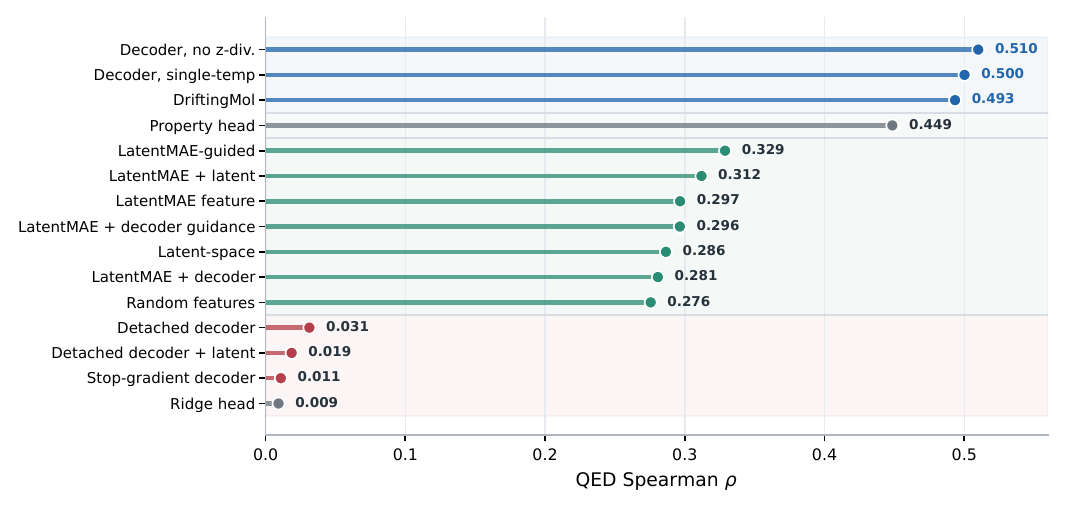}
\caption{Grid-selected QED control across representative mechanism ablations.
The shaded bands separate decoder-coupled models, property-head baselines,
proxy-feature or latent-space controls, and broken-gradient controls. Decoder
features with an open gradient path define the leading control group, while
detached and stop-gradient controls yield near-zero correlation and very low
uniqueness.}
\label{fig:qed-ablation}
\end{figure*}

Table~\ref{tab:qed-main} shows a consistent ordering across the ablation
conditions.
Decoder-coupled settings are the only drifting conditions above $\rho=0.49$.
The default DriftingMol setting achieves $\rho=0.493$ with 94.7\% uniqueness,
while decoder-coupled drift without diversity regularization reaches $\rho=0.510$ but drops to 74.1\%
uniqueness. Table~\ref{tab:qed-diversity} makes the quality profile explicit:
all key conditions retain near-perfect novelty and IntDiv around 0.89, while
DriftingMol and the single-temperature decoder-coupling setting preserve the highest scaffold
diversity. The contrast supports the control-diversity role of
$\mathcal{L}_{\mathrm{zdiv}}$.
Figure~\ref{fig:qed-ablation} gives the corresponding grid-selected view for
the key QED mechanism groups.

\begin{table}[t]
\centering
\caption{Diversity profile for key QED settings at the feasible
$\alpha$ used in Table~\ref{tab:qed-main}. V and N are percentages; IntDiv and
Scaf. Div. are raw diversity scores.}
\label{tab:qed-diversity}
\resizebox{\columnwidth}{!}{%
\begin{tabular}{l c c c c}
\toprule
Model / control & V (\%) & N (\%) & IntDiv & Scaf. Div. \\
\midrule
Decoder coupling (single $\tau$) & 100.0 & 100.0 & 0.895 & 0.581 \\
Decoder coupling (no diversity) & 100.0 & 100.0 & 0.892 & 0.459 \\
DriftingMol & 100.0 & 100.0 & 0.894 & 0.588 \\
Layer-balanced decoder coupling & 100.0 & 99.9 & 0.890 & 0.513 \\
\bottomrule
\end{tabular}
}
\end{table}

External feature quality does not explain the result. The property-enhanced
LatentMAE feature extractor reaches only $\rho=0.329$, despite being trained to
predict molecular properties. Random-feature drift control and latent-space
drift are nearby at 0.276 and 0.286. This clustering suggests that predictive
external features alone are insufficient; the decoder Jacobian is needed to
align the drift gradient with molecular decoding.

\begin{figure}[t]
\centering
\includegraphics[width=\columnwidth]{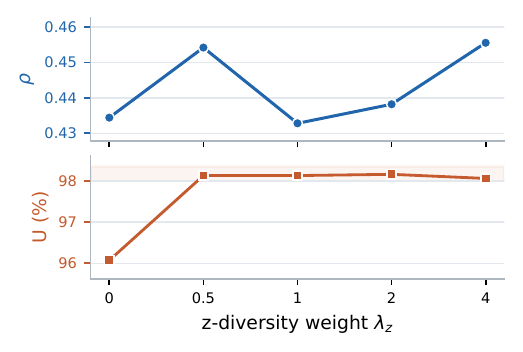}
\caption{Z-diversity sensitivity for the layer-balanced decoder-coupling
setting. The upper trace reports QED control and the lower trace reports
uniqueness; moderate $\lambda_z$ values move the generator into a
high-uniqueness region without eliminating control.}
\label{fig:zdiv-pareto}
\end{figure}

Figure~\ref{fig:zdiv-pareto} summarizes the explicit z-diversity sweep.
Moving from $\lambda_z=0$ to $\lambda_z=0.5$ increases uniqueness from 96.1\%
to 98.1\% and improves QED correlation from 0.434 to 0.454. Larger settings
stay in the same control range ($\rho\approx0.43$--0.46). Thus, in this range,
the diversity term reshapes local sample geometry without replacing the
decoder-feature drift objective.

\begin{table}[t]
\centering
\caption{Three-seed aggregate for key QED settings. Seed 42 uses the reference
configuration; seeds 43--44 provide additional replicates. The 95\% CI is the
t-interval half-width over the three seeds.}
\label{tab:qed-3seed}
\resizebox{\columnwidth}{!}{%
\begin{tabular}{l c c c c}
\toprule
Model / control & Seeds & $n$ & $\rho$ mean & $\rho$ 95\% CI \\
\midrule
Decoder coupling (single $\tau$) & 42,43,44 & 3 & 0.515 & 0.065 \\
Decoder coupling (no diversity) & 42,43,44 & 3 & 0.513 & 0.012 \\
DriftingMol & 42,43,44 & 3 & 0.512 & 0.041 \\
Layer-balanced decoder coupling & 42,43,44 & 3 & 0.445 & 0.018 \\
\bottomrule
\end{tabular}
}
\end{table}

Table~\ref{tab:qed-3seed} separates single-run rankings from reproducibility.
All four key QED settings have three seed replicates. The
single-temperature decoder coupling, diversity-free decoder coupling, and
default DriftingMol settings fall within a similar leading range, with mean
$\rho$ between 0.512 and 0.515, while the layer-balanced decoder-coupling
setting trades some control strength for higher uniqueness.

\begin{figure}[t]
\centering
\includegraphics[width=\columnwidth]{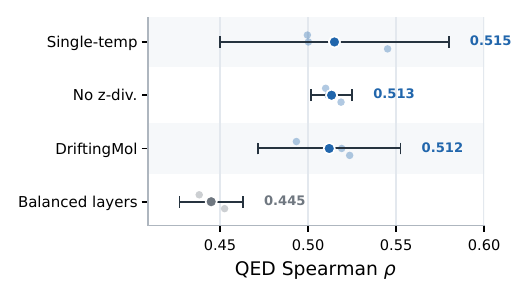}
\caption{Three-seed QED stability. Points show seed means with 95\% confidence
intervals; faint dots show individual seeds.}
\label{fig:qed-seed-ci}
\end{figure}

Figure~\ref{fig:qed-seed-ci} visualizes this seed-level pattern: the three
decoder-coupled QED settings form a tight top group around $\rho\approx0.51$,
whereas the layer-balanced decoder-coupling setting is lower but stable. With only three
seeds, this is limited
seed-level evidence rather than a full uncertainty study, but it reduces the
risk that the main QED tier is driven by one favorable run.

\subsection{Multi-Property Conditioning}

Table~\ref{tab:multi4} reports the matched four-property results used to test
whether decoder coupling transfers beyond QED-only conditioning.

\begin{table}[t]
\centering
\caption{Protocol-matched four-property conditioning under the v2 no-binning setting.
$\bar{\rho}$ is the mean Spearman correlation across QED, SA, LogP, and MolWt;
Lowest U is the lowest property-wise uniqueness percentage.}
\label{tab:multi4}
\resizebox{\columnwidth}{!}{%
\begin{tabular}{l c c c c c c c}
\toprule
Model / control & $\alpha$ & QED & SA & LogP & MolWt & $\bar{\rho}$ & Lowest U (\%) \\
\midrule
Decoder coupling (single $\tau$) & 5.0 & 0.367 & 0.529 & 0.708 & 0.790 & 0.598 & 88.2 \\
Decoder coupling (no diversity) & 5.0 & 0.315 & 0.530 & 0.682 & 0.793 & 0.580 & 72.8 \\
DriftingMol & 5.0 & 0.323 & 0.488 & 0.638 & 0.790 & 0.560 & 91.5 \\
Layer-balanced decoder coupling & 5.0 & 0.252 & 0.378 & 0.561 & 0.705 & 0.474 & 98.6 \\
\midrule
Latent-space drift & 5.0 & 0.188 & 0.271 & 0.496 & 0.552 & 0.377 & 97.9 \\
Random-feature drift control & 5.0 & 0.187 & 0.282 & 0.506 & 0.518 & 0.373 & 97.0 \\
\midrule
No-drift baseline & 1.5 & -0.005 & 0.024 & 0.003 & 0.009 & 0.008 & 98.0 \\
Stop-gradient decoder control & 2.0 & -0.016 & 0.007 & -0.036 & 0.015 & -0.008 & 97.6 \\
\bottomrule
\end{tabular}
}
\end{table}

\begin{figure*}[t]
\centering
\includegraphics[width=0.88\textwidth]{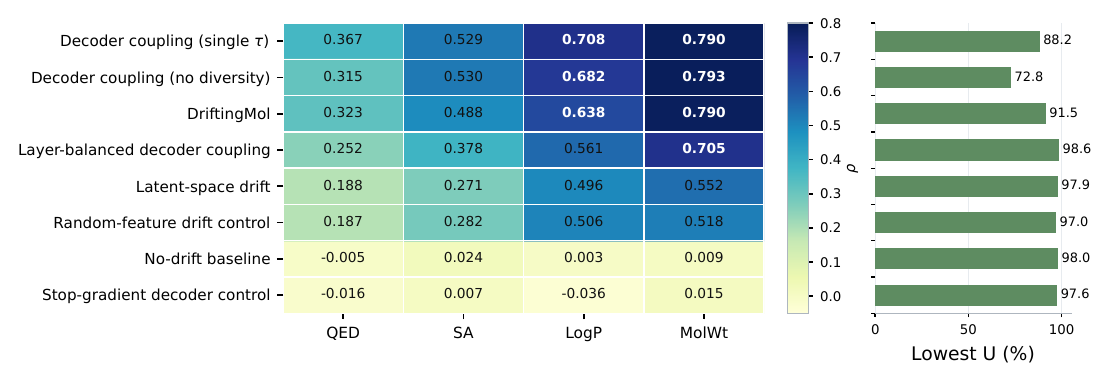}
\caption{Four-property control under the matched v2 no-binning protocol. Decoder-coupled
settings preserve the single-property hierarchy when conditioning on QED, SA,
LogP, and MolWt simultaneously. The right panel reports the minimum
property-wise uniqueness percentage across the four conditioning tasks.}
\label{fig:multi4}
\end{figure*}

Figure~\ref{fig:multi4} visualizes the property-wise correlations and lowest
uniqueness under the matched four-property protocol.
The decoder-coupled advantage transfers to simultaneous conditioning.
The single-temperature decoder-coupling setting reaches $\bar{\rho}=0.598$, the diversity-free
decoder-coupling setting reaches 0.580, and default DriftingMol reaches 0.560 with the best
lowest uniqueness among this group. The layer-balanced decoder-coupling follow-up
reaches $\bar{\rho}=0.474$ with 98.6\% minimum uniqueness. Latent-space and
random-feature controls stay near $\bar{\rho}=0.37$, while no-drift and
stop-gradient controls are near zero.

\section{Analysis}

\paragraph{Decoder coupling is the main explanatory factor.}
The ablation isolates three requirements: a structural decoder feature space,
an open gradient path, and a diversity regularizer. Removing the decoder
feature space lowers QED control by over 40\%. Blocking or detaching the
gradient path yields near-zero QED correlation and uniqueness below 1\%.
Removing z-diversity increases conditional correlation but substantially
reduces uniqueness.

\paragraph{Why trained $\varphi$ is insufficient.}
A trained LatentMAE can encode property-predictive information, but its
Jacobian does not describe how the frozen decoder changes molecules.
Decoder-coupled drift supplies both representation and metric: distances define
reference attraction in decoder space, while
$\partial \varphi_{\mathrm{dec}}/\partial z$ weights generator updates by the
decoder's local structural sensitivity.

\paragraph{Implementation-sensitivity checks.}
Six sensitivity runs test whether simplified implementation choices explain
the effect. Table~\ref{tab:impl-checks} reports these single-seed QED checks.
Property-only 64-reference variants remain below the decoder-coupled group,
whereas the default uses a hybrid-positive, 2048-reference protocol. Removing
CFG weakens property control, removing
z-diversity primarily reduces uniqueness, and removing cross-normalization
substantially reduces property control. Together, these checks indicate that
the effect depends on decoder-coupled drift with the normalization, guidance,
and diversity components used in DriftingMol, rather than on an incidental
implementation choice.

\begin{table}[t]
\centering
\caption{Implementation-sensitivity checks. Values are single-seed QED audit
results; $\rho$ is Spearman correlation and U is uniqueness.}
\label{tab:impl-checks}
\resizebox{\columnwidth}{!}{%
\begin{tabular}{l c c l}
\toprule
Variant & $\rho$ & U (\%) & Interpretation \\
\midrule
Property-only, 64 refs, batch $\lambda_\tau$ & 0.295 & 98.2 & below decoder-coupled group \\
Property-only, 64 refs, fixed $\lambda_\tau$ & 0.298 & 98.5 & below decoder-coupled group \\
No CFG & 0.196 & 98.5 & weaker conditional control \\
No z-diversity & 0.306 & 77.8 & lower uniqueness \\
Y-only normalization & 0.377 & 98.1 & partial control remains \\
No cross-normalization & -0.001 & 89.9 & near-null control \\
\bottomrule
\end{tabular}
}
\end{table}

\paragraph{Architecture sensitivity.}
Alternative SELFIES VAEs preserve the qualitative mechanism but expose its
dependence on latent geometry. Low-beta and latent-128 checkpoints support
downstream QED drifting with $\rho=0.437$ and 0.421, respectively, while a
high-beta VAE and a deeper decoder are weaker ($\rho=0.282$ and 0.272). These
results indicate that the mechanism persists for some alternative VAE settings,
but also depends on VAE latent geometry.

\paragraph{Representation stress test.}
Because SELFIES supplies molecular validity by construction, we keep
representation effects separate from the main mechanism study. The graph-route
stress protocol retrains a graph VAE, rebuilds its latent cache, trains a graph
LatentMAE feature map, and then evaluates QED control, LogP control,
raw-versus-repaired decoding validity, and a no-drift graph ablation under the
same V/U/N and Spearman reporting contract. The completed graph route exposes
the boundary: repaired graph validity can reach 100.0\%, but QED drifting
remains weak ($\rho=0.019$ with 13.1\% uniqueness) and does not outperform the
no-drift graph ablation ($\rho=0.046$). LogP control is stronger
($\rho=0.327$) but remains low-diversity. Raw graph validity also declines
substantially at higher decoding temperatures before repair. We therefore
interpret graph results as representation-sensitivity evidence rather than as a
competing graph-generation result.

\paragraph{Scope.}
DriftingMol is property-biased rather than property-exact. For QED, slopes are
below 1.0 and high targets regress toward the data manifold. This is still
useful for enriching generated libraries toward target regions, but it is not a
replacement for precise molecular optimization. The main advantage is that
generation uses one DiT forward pass plus SELFIES decoding; on an idle RTX
4090D, our benchmark generates 20,000 samples at 47,533 generator
latents/s and 4,604 end-to-end molecules/s.

\section{Limitations and Broader Impact}

Our goal is to isolate the effect of decoder-coupled drift under a fixed
evaluation protocol rather than to compare against all molecular optimization
systems.
External systems such as LIMO, FREED, and MOOD optimize endpoint
properties, run iterative search or denoising, or report distributional metrics
rather than prescribed target-value Spearman/MAE under a one-generator-pass protocol.
They are therefore important references but not protocol-matched baselines for
the present claim. SELFIES validity should also be read as a representation
guarantee, not evidence that the drift objective itself solves chemical
validity. The substantive claims are property control, diversity preservation,
low-cost inference, and ablation evidence for decoder-coupled drift. The
graph-route stress protocol makes this boundary explicit by evaluating raw
graph decoding, repaired graph decoding, and graph latent drifting separately
from the SELFIES main route.

Molecular generation is dual-use. DriftingMol targets coarse drug-like property
conditioning on ZINC250K and does not guarantee synthesizability, biological
activity, or safety. Generated molecules require downstream filtering and
expert review before any experimental use.

\section{Conclusion}

DriftingMol adapts drifting models to SELFIES latent molecular generation and
supports the frozen decoder as an effective decoder-aligned feature space for
drift when its gradient path remains coupled to the generator. Across QED and
four-property conditioning, decoder-coupled settings outperform latent-space,
random-feature, and trained-feature controls while preserving one generator
pass plus one frozen decoder pass at inference. Sensitivity checks show that
this conclusion is not explained by
temperature handling, scalar weighting, CFG alone, or normalization
shortcuts.

\bibliography{references}

@misc{deng2026drifting,
  title = {Generative Modeling via Drifting},
  author = {Deng, Mingyang and Li, He and Li, Tianhong and Du, Yilun and He, Kaiming},
  year = {2026},
  eprint = {2602.04770},
  archivePrefix = {arXiv},
  primaryClass = {cs.LG},
  doi = {10.48550/arXiv.2602.04770},
  url = {https://arxiv.org/abs/2602.04770}
}

@inproceedings{digress,
  title = {{DiGress}: Discrete Denoising Diffusion for Graph Generation},
  author = {Vignac, Clement and Krawczuk, Igor and Siraudin, Antoine and Wang, Bohan and Cevher, Volkan and Frossard, Pascal},
  booktitle = {International Conference on Learning Representations},
  year = {2023},
  eprint = {2209.14734},
  archivePrefix = {arXiv},
  url = {https://openreview.net/forum?id=UaAD-Nu86WX}
}

@inproceedings{moflow,
  title = {{MoFlow}: An Invertible Flow Model for Generating Molecular Graphs},
  author = {Zang, Chengxi and Wang, Fei},
  booktitle = {Proceedings of the 26th ACM SIGKDD International Conference on Knowledge Discovery and Data Mining},
  pages = {617--626},
  year = {2020},
  doi = {10.1145/3394486.3403104}
}

@inproceedings{jtvae,
  title = {Junction Tree Variational Autoencoder for Molecular Graph Generation},
  author = {Jin, Wengong and Barzilay, Regina and Jaakkola, Tommi},
  booktitle = {Proceedings of the 35th International Conference on Machine Learning},
  series = {Proceedings of Machine Learning Research},
  volume = {80},
  pages = {2323--2332},
  publisher = {PMLR},
  year = {2018},
  url = {https://proceedings.mlr.press/v80/jin18a.html}
}

@inproceedings{graphvae,
  title = {{GraphVAE}: Towards Generation of Small Graphs Using Variational Autoencoders},
  author = {Simonovsky, Martin and Komodakis, Nikos},
  booktitle = {Artificial Neural Networks and Machine Learning -- ICANN 2018},
  series = {Lecture Notes in Computer Science},
  pages = {412--422},
  publisher = {Springer},
  year = {2018},
  doi = {10.1007/978-3-030-01418-6_41}
}

@misc{graphnvp,
  title = {{GraphNVP}: An Invertible Flow Model for Generating Molecular Graphs},
  author = {Madhawa, Kaushalya and Ishiguro, Katushiko and Nakago, Kosuke and Abe, Motoki},
  year = {2019},
  eprint = {1905.11600},
  archivePrefix = {arXiv},
  primaryClass = {stat.ML},
  doi = {10.48550/arXiv.1905.11600},
  url = {https://arxiv.org/abs/1905.11600}
}

@inproceedings{gdss,
  title = {Score-based Generative Modeling of Graphs via the System of Stochastic Differential Equations},
  author = {Jo, Jaehyeong and Lee, Seul and Hwang, Sung Ju},
  booktitle = {Proceedings of the 39th International Conference on Machine Learning},
  series = {Proceedings of Machine Learning Research},
  volume = {162},
  pages = {10362--10383},
  publisher = {PMLR},
  year = {2022},
  eprint = {2202.02514},
  archivePrefix = {arXiv},
  url = {https://proceedings.mlr.press/v162/jo22a.html}
}

@article{irwin2012zinc,
  title = {{ZINC}: A Free Tool to Discover Chemistry for Biology},
  author = {Irwin, John J. and Sterling, Teague and Mysinger, Michael M. and Bolstad, Erin S. and Coleman, Ryan G.},
  journal = {Journal of Chemical Information and Modeling},
  volume = {52},
  number = {7},
  pages = {1757--1768},
  year = {2012},
  doi = {10.1021/ci3001277}
}

@article{krenn2020selfies,
  title = {Self-Referencing Embedded Strings ({SELFIES}): A 100\% Robust Molecular String Representation},
  author = {Krenn, Mario and H{\"a}se, Florian and Nigam, AkshatKumar and Friederich, Pascal and Aspuru-Guzik, Al{\'a}n},
  journal = {Machine Learning: Science and Technology},
  volume = {1},
  number = {4},
  pages = {045024},
  year = {2020},
  doi = {10.1088/2632-2153/aba947}
}

@article{krenn2022selfies,
  title = {{SELFIES} and the Future of Molecular String Representations},
  author = {Krenn, Mario and Ai, Qianxiang and Barthel, Senja and Carson, Nessa and Frei, Angelo and Frey, Nathan C. and Friederich, Pascal and Gaudin, Th{\'e}ophile and Gayle, Alberto Alexander and Jablonka, Kevin Maik and Lameiro, Rafael F. and Lemm, Dominik and Lo, Alston and Moosavi, Seyed Mohamad and N{\'a}poles-Duarte, Jos{\'e} Manuel and Nigam, AkshatKumar and Pollice, Robert and Rajan, Kohulan and Schatzschneider, Ulrich and Schwaller, Philippe and Skreta, Marta and Smit, Berend and Strieth-Kalthoff, Felix and Sun, Chong and Tom, Gary and {von Rudorff}, Guido Falk and Wang, Andrew and White, Andrew D. and Young, Adamo and Yu, Rose and Aspuru-Guzik, Al{\'a}n},
  journal = {Patterns},
  volume = {3},
  number = {10},
  pages = {100588},
  year = {2022},
  doi = {10.1016/j.patter.2022.100588}
}

@misc{ho2022classifierfree,
  title = {Classifier-Free Diffusion Guidance},
  author = {Ho, Jonathan and Salimans, Tim},
  howpublished = {arXiv preprint},
  year = {2022},
  eprint = {2207.12598},
  archivePrefix = {arXiv},
  primaryClass = {cs.LG},
  doi = {10.48550/arXiv.2207.12598},
  url = {https://arxiv.org/abs/2207.12598}
}

@inproceedings{peebles2023scalable,
  title = {Scalable Diffusion Models with Transformers},
  author = {Peebles, William and Xie, Saining},
  booktitle = {2023 IEEE/CVF International Conference on Computer Vision},
  pages = {4172--4182},
  year = {2023},
  doi = {10.1109/ICCV51070.2023.00387}
}

@article{gomez2018automatic,
  title = {Automatic Chemical Design Using a Data-Driven Continuous Representation of Molecules},
  author = {G{\'o}mez-Bombarelli, Rafael and Wei, Jennifer N. and Duvenaud, David and Hern{\'a}ndez-Lobato, Jos{\'e} Miguel and S{\'a}nchez-Lengeling, Benjam{\'i}n and Sheberla, Dennis and Aguilera-Iparraguirre, Jorge and Hirzel, Timothy D. and Adams, Ryan P. and Aspuru-Guzik, Al{\'a}n},
  journal = {ACS Central Science},
  volume = {4},
  number = {2},
  pages = {268--276},
  year = {2018},
  doi = {10.1021/acscentsci.7b00572}
}

@article{olivecrona2017molecular,
  title = {Molecular De-novo Design through Deep Reinforcement Learning},
  author = {Olivecrona, Marcus and Blaschke, Thomas and Engkvist, Ola and Chen, Hongming},
  journal = {Journal of Cheminformatics},
  volume = {9},
  number = {1},
  pages = {48},
  year = {2017},
  doi = {10.1186/s13321-017-0235-x}
}

@inproceedings{eckmann2022limo,
  title = {{LIMO}: Latent Inceptionism for Targeted Molecule Generation},
  author = {Eckmann, Peter and Sun, Kunyang and Zhao, Bo and Feng, Mudong and Gilson, Michael and Yu, Rose},
  booktitle = {Proceedings of the 39th International Conference on Machine Learning},
  series = {Proceedings of Machine Learning Research},
  volume = {162},
  pages = {5777--5792},
  publisher = {PMLR},
  year = {2022},
  url = {https://proceedings.mlr.press/v162/eckmann22a.html}
}

@article{zhou2019optimization,
  title = {Optimization of Molecules via Deep Reinforcement Learning},
  author = {Zhou, Zhenpeng and Kearnes, Steven and Li, Li and Zare, Richard N. and Riley, Patrick},
  journal = {Scientific Reports},
  volume = {9},
  number = {1},
  pages = {10752},
  year = {2019},
  doi = {10.1038/s41598-019-47148-x}
}

@inproceedings{yang2021freed,
  title = {Hit and Lead Discovery with Explorative {RL} and Fragment-based Molecule Generation},
  author = {Yang, Soojung and Hwang, Doyeong and Lee, Seul and Ryu, Seongok and Hwang, Sung Ju},
  booktitle = {Advances in Neural Information Processing Systems},
  volume = {34},
  pages = {7924--7936},
  publisher = {Curran Associates, Inc.},
  year = {2021},
  url = {https://papers.nips.cc/paper_files/paper/2021/hash/41da609c519d77b29be442f8c1105647-Abstract.html}
}

@inproceedings{lee2023mood,
  title = {Exploring Chemical Space with Score-based Out-of-distribution Generation},
  author = {Lee, Seul and Jo, Jaehyeong and Hwang, Sung Ju},
  booktitle = {Proceedings of the 40th International Conference on Machine Learning},
  series = {Proceedings of Machine Learning Research},
  volume = {202},
  pages = {18872--18892},
  publisher = {PMLR},
  year = {2023},
  url = {https://proceedings.mlr.press/v202/lee23f.html}
}

@inproceedings{lipman2023flow,
  title = {Flow Matching for Generative Modeling},
  author = {Lipman, Yaron and Chen, Ricky T. Q. and Ben-Hamu, Heli and Nickel, Maximilian and Le, Matthew},
  booktitle = {International Conference on Learning Representations},
  year = {2023},
  eprint = {2210.02747},
  archivePrefix = {arXiv},
  primaryClass = {cs.LG},
  url = {https://openreview.net/forum?id=PqvMRDCJT9t}
}

\end{document}